\documentclass[letterpaper, 10 pt, conference]{ieeeconf}  

\IEEEoverridecommandlockouts                              

\usepackage{graphics} 
\usepackage{amsmath} 
\usepackage{amssymb}  
\usepackage[makeroom]{cancel}
\usepackage{colortbl}
\usepackage{pifont}
\usepackage{mdframed}
\usepackage{balance}
\newcommand{\hide}[1]{}

\newcommand{\bdmath}{\begin{dmath}}
\newcommand{\edmath}{\end{dmath}}
\newcommand{\beq}{\begin{equation}}
\newcommand{\eeq}{\end{equation}}
\newcommand{\bdm}{\begin{displaymath}}
\newcommand{\edm}{\end{displaymath}}
\newcommand{\bea}{\begin{eqnarray}}
\newcommand{\eea}{\end{eqnarray}}
\newcommand{\beal}{\beq \begin{array}{ll}}
\newcommand{\eeal}{\end{array} \eeq}
\newcommand{\beas}{\begin{eqnarray*}}
\newcommand{\eeas}{\end{eqnarray*}}
\newcommand{\ba}{\begin{array}}
\newcommand{\ea}{\end{array}}
\newcommand{\bit}{\begin{itemize}}
\newcommand{\eit}{\end{itemize}}
\newcommand{\ben}{\begin{enumerate}}
\newcommand{\een}{\end{enumerate}}

\newcommand{\SO}{\mathrm{SO}}

\newcommand{\SOthree}{\ensuremath{\SO(3)}\xspace}

\newcommand{\R}{\mathtt{R}}

\newcommand{\vel}{\mathbf{v}}

\newcommand{\bias}{\mathbf{b}}

\newcommand{\gravity}{\mathbf{g}}
\newcommand{\noise}{\boldsymbol\eta}

\newcommand{\expmap}{\mathrm{Exp}}

\usepackage[dvipsnames]{xcolor}

\usepackage{amsmath}
\usepackage{amssymb}
\usepackage{amsthm}
\usepackage{mathtools}
\usepackage{xspace}

\usepackage{graphicx}
\usepackage{subcaption}
\usepackage[percent]{overpic}
\usepackage{booktabs}
\usepackage{tabularray}
\UseTblrLibrary{booktabs}
\usepackage{rotating}
\usepackage{tikz}
\usetikzlibrary{arrows.meta}

\usepackage{ragged2e} 
\usepackage{pifont}
\usepackage{lipsum}
\usepackage{gensymb}
\usepackage{comment}
\usepackage{multirow}
\usepackage{tabularray}
\usepackage[export]{adjustbox}
\usepackage{rotating}
\usepackage{makecell}
\usepackage{tikz}

\usetikzlibrary{positioning,arrows.meta,calc, fit}
\usepackage{algorithm}
\usepackage{algpseudocode}

\AtEndPreamble{
    \usepackage[capitalize]{cleveref}
    \crefname{section}{Sec.}{Secs.}
    \Crefname{section}{Section}{Sections}
    \Crefname{table}{Table}{Tables}
    \crefname{table}{Tab.}{Tabs.}
}

\definecolor{darkgreen}{rgb}{0.0,0.6,0.0}

\usepackage{amssymb} 
\usepackage{breqn}
\usepackage{multirow}
\usepackage{cancel} 
\usepackage{xfrac}
\usepackage{xspace}
\usepackage{booktabs}
\usepackage{pifont}
\usepackage{arydshln}
\usepackage{makecell}  
\usepackage{balance}
\usepackage{verbatim}
\usepackage{comment}
\makeatletter
\let\NAT@parse\undefined
\makeatother
\usepackage[colorlinks=true,
			linkcolor=blue,
			urlcolor=blue,
			citecolor=blue]{hyperref}

\makeatletter
\DeclareRobustCommand\onedot{\futurelet\@let@token\@onedot}
\def\@onedot{\ifx\@let@token.\else.\null\fi\xspace}

\makeatother

\definecolor{pastelred}{RGB}{235,119,119}
\definecolor{pastelorange}{RGB}{255,202,126}

\makeatletter
\newcommand{\setword}[2]{%
  \phantomsection
  #1\def\@currentlabel{\unexpanded{#1}}\label{#2}%
}
\makeatother

\title{\LARGE \bf
DefVINS: Visual-Inertial Odometry for Deformable Scenes
}

\author{Samuel Cerezo$^{1}$ and Javier Civera$^{1}$%
\thanks{$^{1}$Authors are with the I3A, Universidad de Zaragoza, Spain. {\tt\small \{samueladriancerezo, jcivera\}@unizar.es}}%
}

\begin{document}

\maketitle
\thispagestyle{empty}
\pagestyle{empty}

\begin{abstract}
Deformable scenes violate the rigidity assumption underlying classical
visual-inertial odometry (VIO), causing conventional methods to either overfit local non-rigid motion or suffer severe camera-pose drift when scene deformation dominates visual parallax.
In this paper, we introduce DefVINS, the first VIO pipeline specifically designed for deformable environments. DefVINS decomposes the odometry state into camera motion components, anchored by inertial measurements, and a non-rigid scene deformation modeled through an embedded deformation graph.
We further introduce VIMandala, the first benchmark for VIO in deformable scenes, featuring real visual-inertial sequences with ground-truth camera trajectories. To enable controlled evaluation, we additionally augment the synthetic Drunkard's benchmark with simulated inertial measurements.
On the theoretical side, we present an observability analysis of deformable VIO, characterizing how inertial measurements constrain camera motion and make otherwise unobservable modes identifiable in the presence of scene deformation. 
Experiments on both the synthetic Drunkard's benchmark and the real VIMandala benchmark demonstrate that DefVINS consistently outperforms rigid visual-inertial and non-rigid visual-only odometry baselines. Source code and data will be released upon acceptance.
\end{abstract}

\section{Introduction}
Visual-Inertial Odometry (VIO) is a foundational technology for modern mobile robotics and augmented, mixed and virtual reality~\cite{cadena2016past}. 
Among the different sensing modalities, the combination of cameras and Inertial Measurement Units (IMUs) stands out as a particularly robust and complementary source of motion information. The IMU delivers high-frequency measurements of angular velocity and linear acceleration, enabling reliable short-term motion estimation under challenging visual conditions such as low texture or rapid motion. Moreover, inertial measurements directly constrain roll, pitch and metric scale, which are unobservable from monocular vision alone~\cite{Huang2019VisualInertialNA}. Conversely, cameras provide exteroceptive measurements of the environment through visual landmarks, constraining camera motion over longer temporal windows and compensating for the drift accumulated by inertial integration~\cite{usenko2016direct, VINSmono}. This complementary fusion has enabled VIO systems, such as VINS-Mono~\cite{VINSmono} and OKVIS~\cite{leutenegger2015keyframe}, to achieve highly accurate and robust ego-motion estimation in rigid environments, establishing VIO as a mature and widely adopted technology~\cite{krishnan2025benchmarking}.

However, the success of current VIO methods fundamentally relies on the assumption that the observed scene is rigid. This assumption is frequently violated in real-world scenarios involving deformable objects and surfaces, such as human bodies, clothing, or soft materials. When non-rigid scene motion is not explicitly modeled, the geometric assumptions underlying VIO break down, leading to significant degradation in pose estimation. In particular, the estimator may initially overfit the rigid motion model to local deformations, incorrectly attributing non-rigid scene motion to camera motion. More severely, when deformation-induced parallax dominates the apparent inter-frame motion, the visual measurements may provide insufficient constraints on the true camera motion, resulting in substantial pose drift. Consequently, robust and accurate localization in highly deformable environments remains a fundamental open challenge for visual-only and visual-inertial perception~\cite{challenges2022}.

Beyond the challenges of explicitly modeling non-rigid scene motion, moving from rigid to deformable environments fundamentally changes the structure of the underlying estimation problem. Classical analyses of VIO observability have established that inertial sensing renders scale and gravity observable in rigid scenes. How these observability properties extend to deformable environments, however, remains largely unexplored. Scene deformation introduces additional latent degrees of freedom that may couple with camera motion, giving rise to ambiguities and ill-conditioned estimation problems when relying on visual measurements alone. In this setting, inertial measurements provide complementary constraints by anchoring the rigid component of camera motion to its underlying short-term dynamics, thereby restricting the solution space and reducing spurious couplings between ego-motion and non-rigid deformation. Building on this observation, we present an explicit observability analysis for the deformable visual-inertial odometry estimation case. Our analysis characterizes the structural constraints introduced by inertial sensing and reveals how they eliminate otherwise unobservable modes and improve the conditioning compared with purely visual formulations.

In this paper, we introduce DefVINS, the first VIO pipeline explicitly designed for deformable scenes. DefVINS incorporates an embedded deformation graph to model the non-rigid motion of the scene, effectively decoupling the rigid, IMU-anchored state from the scene’s non-rigid degrees of freedom.
Given the absence of benchmarks specifically targeting non-rigid VIO, we also contribute two new evaluation resources: (1) an extension of the synthetic Drunkard’s dataset augmented with inertial measurements, and (2) VIMandala, a novel real-world dataset featuring visual-inertial data and camera pose ground truth while observing a deformable mandala cloth.
Experimental results on both datasets demonstrate that DefVINS consistently outperforms rigid visual-inertial methods as well as existing deformable visual-only odometry baselines.

\section{Related Work}

Visual-inertial fusion has been extensively studied for camera motion and scene-structure estimation, mainly using feature-based~\cite{leutenegger2015keyframe,VINSmono,campos2021orb}, direct~\cite{concha2016visual,von2018direct}, and learning-based approaches~\cite{clark2017vinet,peng2024dvi,lozano2026vi3}.
The observability and consistency of VIO have also been thoroughly investigated under rigid-scene assumptions, establishing the conditions under which quantities such as scale, gravity, initial velocity, and IMU biases become observable~\cite{Hesch2014Observability,Martinelli2013ClosedForm,huang2009observability}. These analyses, however, assume that the observed scene geometry is rigid, and therefore do not account for the additional degrees of freedom introduced by non-rigid scene motion.

When the rigidity assumption does not hold, some odometry and SLAM methods address dynamic scenes by identifying and suppressing independently moving objects through semantic segmentation or motion-consistency tests~\cite{bescos2018dynaSLAM,song2022dynavins,song2024dynavins++}.
These approaches can effectively mitigate moderate dynamic content while preserving the rigid scene structure required for camera-motion estimation. However, they become inherently limited when non-rigid motion dominates the observations or when few rigid scene elements remain. An alternative class of methods explicitly tracks dynamic objects and jointly estimates their motion~\cite{ballester2021dot,bescos2021dynaslam,judd2024multimotion,morris2025dynosam} with the camera's one. Although more tolerant to the absence of a static background, these methods generally represent the scene as a collection of independently moving rigid bodies and therefore do not capture continuous non-rigid deformation.

A complementary body of work explicitly models non-rigid geometry, with applications including human and medical reconstruction. Methods such as DynamicFusion~\cite{Newcombe2015DynamicFusion}, DefSLAM~\cite{Lamarca2020DefSLAM}, and NR-SLAM~\cite{gomez2024nrslam} represent deformation through embedded deformation graphs or learned warp fields, enabling the reconstruction and tracking of substantially deforming scenes. However, these approaches only rely on RGB or RGB-D observations and do not exploit inertial sensors, despite modern IMUs being lightweight, inexpensive, compact, and ubiquitous in mobile devices. In these setups, camera motion and scene deformation can become strongly coupled, particularly when deformation-induced parallax dominates the observed motion. This coupling introduces ambiguities and can lead to ill-conditioned pose estimation.

From a state-estimation perspective, observability has been extensively analyzed for rigid VIO systems~\cite{Hesch2014Observability,huang2009observability,huai2022observability}. Deformable scenes fundamentally alter this structure: part of the observed parallax can be explained either by camera motion or by scene deformation, introducing additional ambiguities that are not captured by existing rigid-scene analyses. In particular, deformation can effectively absorb visual evidence of camera motion, while the resulting coupling depends on the magnitude and spatial structure of the deformation as well as on the available inertial excitation. To the best of our knowledge, no previous work explicitly characterizes the observability and conditioning of visual-inertial odometry in the presence of non-rigid deformation, nor analyzes how inertial measurements constrain the resulting coupled estimation problem.

To the best of our knowledge, no existing method jointly
(i) integrates an explicit deformation model within a visual-inertial odometry
pipeline, and
(ii) maintains a rigid, IMU-anchored reference to ensure metric consistency.
Our work addresses the above described gaps by jointly modeling camera motion and non-rigid scene deformation within a visual-inertial odometry framework. DefVINS represents deformation using an embedded deformation graph while maintaining an IMU-anchored rigid motion component, providing metric and dynamical constraints that are absent from purely visual deformable odometry. In addition, we provide an explicit observability analysis of the resulting estimation problem and show how inertial constraints reduce the coupling between camera motion and deformation, improving the conditioning of the optimization under strong non-rigid motion.

\section{Visual-Inertial Odometry}

\subsection{State definition}
\label{sec:state-definition}

Our state is defined over a set of $N$ consecutive keyframes in a sliding window that we denote by their indices $\mathcal{W} = \{ \tau \}_{\tau=1}^{N}$. 
For every keyframe at a specific time instant $t_\tau$, $\tau \in \mathcal{W}$, we store its orientation $\R_\tau \in \SOthree$,
velocity $\mathbf{v}_\tau \in \mathbb{R}^3$ and position $\mathbf{t}_\tau \in \mathbb{R}^3$ in a global reference frame. We consider short sequences and therefore assume constant gyroscope and accelerometer biases $\mathbf{b}^g,
\mathbf{b}^a \in \mathbb{R}^3$. The direction of gravity is represented as $\hat{\mathbf{g}} = \frac{\mathbf{g}}{||\mathbf{g}||} \in \mathbb{S}^2$. 
To account for non-rigid scene dynamics, the state is augmented with
$\{ \mathbf{x}_i^\tau \}_{i \in \mathcal{D},\tau \in \mathcal{W}}$, which collects the positions of the set $\mathcal{D} = \{ i \}_{i=1}^{D}$ of $D$ deformation nodes active within
$\mathcal{W}$. The full state stacks all these variables as a single vector
\begin{equation}
\label{eq:obs-vector-state-vio}
\small
\boldsymbol{\xi}
=
\mathrm{stack}\Big(
\{\R_{\tau}, \mathbf{v}_{\tau}, \mathbf{t}_{\tau}\}_{\tau \in \mathcal{W}},\;
\mathbf{b}^g, 
\mathbf{b}^a,
\hat{\mathbf{g}},\;
\{ \mathbf{x}_i^\tau \}_{i \in \mathcal{D},\tau \in \mathcal{W}}
\Big)
\end{equation}

\subsection{IMU preintegration and gravity residuals}

The IMU kinematics between two consecutive keyframes, taken at time instants $t_\tau$ and $t_{\tau+1}$, are as follows
\begin{align}
\R_{\tau+1}
&= \R_\tau
\prod_{t_k \in \left[ t_\tau, t_{\tau+1} \right)}
\expmap\!\left(
\left(\tilde{\boldsymbol{\omega}}_k - \bias^g - \noise^{g}_k\right)\,\Delta t
\right),  \label{eq:preintegrationRotation}\\
\vel_{\tau+1}
&= \vel_\tau + \gravity\,\Delta T
+ \sum_{t_k \in \left[ t_\tau, t_{\tau+1} \right)}
\R_k \big(\tilde{\mathbf{a}}_k - \bias^a - \noise^{a}_k\big)\,\Delta t,
\label{eq:preintegrationVelocity}\\
\mathbf{t}_{\tau+1}
&= \mathbf{t}_\tau
+ \sum_{t_k \in \left[ t_\tau, t_{\tau+1} \right)}
\left(
\vel_k \Delta t
+ \tfrac{1}{2}\gravity\,\Delta t^2
\right) \nonumber \\
&\quad + \sum_{t_k \in \left[ t_\tau, t_{\tau+1} \right)}
\frac{1}{2} \R_k \left(\tilde{\mathbf{a}}_k - \bias^a - \noise^{a}_k\right)\,\Delta t^2, 
\label{eq:preintegrationTranslation}
\end{align}

\noindent
where $\Delta t$ is the IMU sampling period and $\tilde{\boldsymbol{\omega}}_k$ and $\tilde{\mathbf{a}}_k$ correspond to the
gyroscope and accelerometer readings at $t_k \in \left[ t_\tau, t_{\tau+1}\right)$, which are affected by biases
$\bias^g$, $\bias^a$ and additive noise $\noise^{g}_k$, $\noise^{a}_k$.
Between time instants $t_\tau$ and $t_{\tau+1}$, the relative motion is constrained by rotation, velocity, and position residuals,
\begin{align}
\mathbf{r}_{\Delta \R} &= 
\mathrm{Log}\left( 
\Delta \tilde{\R}_{\tau,\tau+1}^{\top} \R_{\tau}^{\top} \R_{\tau+1}
\right) \\
\mathbf{r}_{\Delta \mathbf{v}} &=
\R_\tau^{\top}(
\mathbf{v}_{\tau+1} - \mathbf{v}_{\tau} - \mathbf{g}\Delta T) - \Delta \tilde{\mathbf{v}}_{\tau,\tau+1} \\
\mathbf{r}_{\Delta \mathbf{t}}&=\R_\tau^{\top}(
\mathbf{t}_{\tau+1} - \mathbf{t}_{\tau} - \mathbf{v}_{\tau} \Delta T
- \tfrac{1}{2}\mathbf{g}\Delta T^2) - \Delta \tilde{\mathbf{t}}_{\tau,\tau+1}
\end{align}
\noindent where $\Delta\tilde{\R}_{\tau,\tau+1}$, $\Delta\tilde{\mathbf{v}}_{\tau,\tau+1}$ and $\Delta\tilde{\mathbf{t}}_{\tau,\tau+1}$ are preintegrated from the IMU readings~\cite{On-manifold2017}.
Each of them is weighted by its corresponding covariance matrix ${{\boldsymbol\Sigma}_{\Delta \boldsymbol\phi}}$, ${\boldsymbol\Sigma}_{\Delta \mathbf{v}}$ and ${\boldsymbol\Sigma}_{\Delta \mathbf{t}}$ obtained during the preintegration process as described in~\cite{cerezoGNSS}, Sec.~III--A.
Additionally, a gravity residual is introduced to enforce consistency between the estimated accelerations and the gravity direction,
\begin{equation}
\label{eq:gravity-residual}
\mathbf{r}_{\mathbf{g}} =
\left(
\frac{\mathbf{v}_{\tau+1} - \mathbf{v}_{\tau}}{\Delta T}
-
\frac{\R_{\tau} \Delta \mathbf{v}_{\tau,\tau+1}}{\Delta T}
\right)
-
\|\mathbf{g}\|\,\hat{\mathbf{g}},
\end{equation}
where $\hat{\mathbf{g}} \in \mathbb{S}^2$ and $\|\mathbf{g}\| = 9.81~\mathrm{m/s}^2$.
The contribution of the aforementioned residuals to the overall cost function can be aggregated as 
\begin{equation}
\mathcal{L}_{\mathrm{imu}}^\tau
=
\|\mathbf{r}_{\Delta \R}\|_{{\boldsymbol\Sigma}_{\Delta \boldsymbol\phi}}^{}
+
\|\mathbf{r}_{\Delta \mathbf{v}}\|_{{\boldsymbol\Sigma}_{\Delta \mathbf{v}}}^{}
+
\|\mathbf{r}_{\Delta \mathbf{t}}\|_{{\boldsymbol\Sigma}_{\Delta \mathbf{t}}}^{}
+
\|\mathbf{r}_{\mathbf{g}}\|_{{\boldsymbol\Sigma}_{\mathbf{g}}}^{}
\label{eq:imu-term}
\end{equation}

\subsection{Reprojection error}

Each tracked feature will contribute with its reprojection error, that will be aggregated for the keyframe at time $t_\tau$ as
\begin{equation}
\label{eq:vision-term}
\mathcal{L}_{\mathrm{rep}}^{\tau}
=
\sum_{i \in \mathcal{P}}
\left\|
\mathbf{z}_{i}^\tau
-
\pi\!\left(\R_\tau,\mathbf{t}_{\tau},\mathbf{x}_{i}^\tau\right)
\right\|_{\boldsymbol{\Sigma}_{i}^\tau}^{},
\end{equation}
where $\mathbf{z}_{i}^\tau \in \Omega$, $\Omega$ being the image domain, denotes the image measurement of the point feature $\mathbf{x}_{i}^\tau \in \mathbb{R}^3$ at time $t_\tau$, $\boldsymbol{\Sigma}_{i}^\tau$ is its associated covariance and $\pi(\cdot)$ the camera projection model. We will denote as $\mathcal{P} = \{ i \}_{i=1}^{M}$ the full set of 3D points tracked. Note that, differently from rigid visual-inertial odometry, the 3D scene points $\mathbf{x}_{i}^\tau$ may change their position at each temporal instant $t_\tau$.

\subsection{Non-rigid residual}
We will assume that the imaged scenes may exhibit non-rigid motions, such as those exhibited by deformable clothes or cables under physical interaction.
We will model these effects by a deformation graph that we build over a subset $\mathcal{D} = \{ i \}_{i=1}^{D} \subset \mathcal{P}$ composed by the $D$ longer feature tracks from $\mathcal{P}$. 
The tracked points $\mathbf{x}_i^{\tau}$ define the nodes of the graph. We will define the reference keyframe as the one taken at $\tau=1$.
Two nodes $i,j \in \mathcal{D}$ are connected by an edge $(i,j)$ if their distance in the reference keyframe is below a spatial threshold.
This provides a compact neighborhood structure that encodes how nearby points of the scene relate to each other.
The reference distance between nodes is defined in the reference keyframe as
\begin{equation}
d_{ij}^{1} = \bigl\lVert \mathbf{x}_i^{1} - \mathbf{x}_j^{1} \bigr\rVert .
\end{equation}

The full non-rigid residual is built up by several terms that we detail next.

\begin{figure*}[]
\scriptsize
\centering
\begin{equation*}
\label{eq:obs-matrix-full-vio}
\mathcal{O}=\left[
\begin{array}{cccccccccc}

-\mathbf{J}_r^{-1}(\boldsymbol{\phi}) \R_{\tau+1}^\top \R_{\tau}
& \mathbf{0}
& \mathbf{0}
& \mathbf{J}_r^{-1}(\boldsymbol{\phi})
& \mathbf{0}
& \mathbf{0}
& \mathbf{0}
& \mathbf{0}
& \mathbf{0}
& \mathbf{0}
\\[1.0ex]

\left( \R_\tau^\top (\mathbf{v}_{\tau+1} - \mathbf{v}_\tau - \mathbf{g}\Delta T) \right)^\wedge
& -\R_\tau^\top
& \mathbf{0}
& \mathbf{0}
&  \R_\tau^\top
& \mathbf{0}
& \mathbf{0}
& -\R_\tau^\top \Delta T
& -\|\mathbf{g}\|\R_\tau^\top \Delta T
& \mathbf{0}
\\[1.0ex]

\left( \R_\tau^\top (\mathbf{t}_{\tau+1} - \mathbf{t}_\tau - \mathbf{v}_\tau\Delta T - \tfrac12\mathbf{g}\Delta T^2) \right)^\wedge
& -\R_\tau^\top\Delta T
& -\R_\tau^\top
& \mathbf{0}
& \mathbf{0}
& \R_\tau^\top
& \mathbf{0}
& -\R_\tau^\top \tfrac12\Delta T^2
& -\tfrac12\|\mathbf{g}\|\R_\tau^\top\Delta T^2
& \mathbf{0}
\\[1.0ex]

\mathbf{H}_{\R_\tau}^{\text{vis}}
& \mathbf{0}
& \mathbf{H}_{\mathbf{t}_\tau}^{\text{vis}}
& \mathbf{H}_{\R_{\tau+1}}^{\text{vis}}
& \mathbf{0}
& \mathbf{H}_{\mathbf{t}_{\tau+1}}^{\text{vis}}
& \mathbf{0}
& \mathbf{0}
& \mathbf{0}
& \mathbf{H}_{\mathrm{nr}}^{\text{vis}}
\\[1.0ex]

\mathbf{0}
& \mathbf{0}
& \mathbf{0}
& \mathbf{0}
& \mathbf{0}
& \mathbf{0}
& \mathbf{0}
& \mathbf{0}
& \mathbf{0}
& \mathbf{H}_{\mathrm{nr}}^{\text{elas}}
\\[1.0ex]

\mathbf{0}
& \mathbf{0}
& \mathbf{0}
& \mathbf{0}
& \mathbf{0}
& \mathbf{0}
& \mathbf{0}
& \mathbf{0}
& \mathbf{0}
& \mathbf{H}_{\mathrm{nr}}^{\text{visc}}
\\[1.0ex]

\mathbf{H}_{\R_\tau}^{\text{photo}}
& \mathbf{0}
& \mathbf{H}_{\mathbf{t}_\tau}^{\text{photo}}
& \mathbf{H}_{\R_{\tau+1}}^{\text{photo}}
& \mathbf{0}
& \mathbf{H}_{\mathbf{t}_{\tau+1}}^{\text{photo}}
& \mathbf{0}
& \mathbf{0}
& \mathbf{0}
& \mathbf{H}_{\mathrm{nr}}^{\text{photo}}
\\[1.0ex]

\mathbf{0}
& \mathbf{0}
& \mathbf{0}
& \mathbf{0}
& \mathbf{0}
& \mathbf{0}
& \mathbf{I}_3
& \mathbf{0}
& \mathbf{0}
& \mathbf{0}
\\[1.0ex]

\mathbf{0}
& \mathbf{0}
& \mathbf{0}
& \mathbf{0}
& \mathbf{0}
& \mathbf{0}
& \mathbf{0}
& \mathbf{I}_3
& \mathbf{0}
& \mathbf{0}
\\[1.0ex]

\mathbf{0}
& -\tfrac{1}{\Delta T}\mathbf{I}_3
& \mathbf{0}
& \mathbf{0}
& \tfrac{1}{\Delta T}\mathbf{I}_3
& \mathbf{0}
& \mathbf{0}
& \mathbf{0}
& -\|\mathbf{g}\|\,\mathbf{J}_{\mathbb{S}^2}
& \mathbf{0}
\end{array}
\right]
\end{equation*}
\caption{\textbf{Structure of the full observability matrix $\mathcal{O}$.}
Each block row corresponds to the Jacobians of the different measurement and constraint terms, including IMU preintegration factors, visual residuals, non--rigid motion priors, and bias and gravity constraints. This structured formulation highlights the contribution of each sensing modality to the overall system observability.}

\label{fig:obs-matrix-explicit-vio}
\end{figure*}

\textbf{Elastic constraint. }We will assume an elastic relation between points in the deformation graph, preventing unrealistic stretching or compression.
If part of the deformable object bends, nearby nodes may also experience non-rigid deformation, but their spacing should not deviate excessively from the reference configuration.
We promote this behavior by penalizing changes in pairwise distances.
\begin{equation}
\mathcal{L}_{ij,\mathrm{elas}}^{\tau}
=
\kappa \,
\frac{\left(d_{ij}^{\tau} - d_{ij}^{1}\right)^{2}}{d_{ij}^{1}},
\qquad
d_{ij}^{\tau}
=
\bigl\lVert \mathbf{x}_i^{\tau} - \mathbf{x}_j^{\tau} \bigr\rVert .
\label{eq:elastic_ral}
\end{equation}

\noindent where $\kappa$ is an elastic deformation constant. In words, this term will penalize excessive stretching or compression, so small deformations will be preferred for the same IMU and multi-view constraints.

\textbf{Viscous constraint. }While the elastic term controls shape deviations between times $t_\tau$ and $t_1$, we additionally impose constraints on the kinematics via a viscous term.
Let $\mathbf{s}_i^{\tau}=\mathbf{x}_i^{\tau}-\mathbf{x}_i^{\tau-1}$
be the displacement of node $i$ between two consecutive keyframes.
Following~\cite{gomez2024nrslam}, we encourage nearby nodes to move in a similar manner
\begin{equation}
\mathcal{L}_{ij,\mathrm{visc}}^{\tau}
=
b_{ij}
\,
\bigl\lVert
\mathbf{s}_i^{\tau} - \mathbf{s}_j^{\tau}
\bigr\rVert^{2},
\label{eq:viscous_ral}
\end{equation}
where proximity is encoded by spatially decaying weights $b_{ij}$, which impose strong temporal coupling between spatially close frames, promoting smooth deformations
\begin{equation}
b_{ij}
=
\exp\!\left(
-\frac{\left\lVert \mathbf{x}_i^{1} - \mathbf{x}_j^{1} \right\rVert^{2}}
     {2\sigma^{2}}
\right),
\label{eq:bij_ral}
\end{equation}

\noindent the constant $\sigma$ modeling the strength of viscous constraints. 

\textbf{Photometric constraint. }In addition to geometric reprojection error, we also leverage image intensity information.
A semi-direct strategy is adopted, in which photometric data association is carried out on Shi–Tomasi features using the modified multi-scale Lucas–Kanade tracker introduced in~\cite{gomez2021sd}.
For every deformation node, its image projections in consecutive keyframes $\tau{-}1$ and $\tau$ are expected to correspond to pixels with similar intensities, following the classical brightness constancy assumption:
\begin{equation}
\mathcal{L}_{i,\mathrm{photo}}^{\tau}
=
\Big(
I^{\tau}(\mathbf{u}_i^{\tau})
-\alpha_i
I^{\tau-1}(\mathbf{u}_i^{\tau-1}) + \beta_i
\Big)^{2},
\label{eq:photo_node}
\end{equation}
\noindent where $\mathbf{u}_i^{\tau}=\pi\left(\R_\tau,\mathbf{t}_{\tau},\mathbf{x}_{i}^\tau\right)$ and $I^{\tau}(\mathbf{u}_i)$ is evaluated via bilinear sampling. Local illumination invariance is achieved by computing local gain $\alpha_i$ and bias $\beta_i$ terms for each tracked image point. 
This term encourages nodes to move consistently with the apparent texture motion in the images.


The combination of the elastic, viscous and photometric terms yields the full non-rigid residual for a keyframe pair $(\tau{-}1,\tau)$:
\begin{equation}
\mathcal{L}_{\mathrm{nr}}^{\tau}
=
\sum_{(i,j)\in\mathcal{E}}
\!\Big(
\mathcal{L}_{ij,\mathrm{elas}}^{\tau}
+
\mathcal{L}_{ij,\mathrm{visc}}^{\tau}
\Big)
+
\sum_{i\in\mathcal{D}}
\mathcal{L}_{i,\mathrm{photo}}^{\tau}
\label{eq:enr_total}
\end{equation}
\noindent where $\mathcal{E}$ is the set of all edges in the deformation graph.


\subsection{Overall cost function}
The visual-inertial estimation problem is formulated as a nonlinear least-squares optimization
over a fixed-size sliding window $\mathcal{W}$.
The window contains the most recent $N$ keyframes and defines the state vector $\xi$
over which constraints are applied.
Within $\mathcal{W}$, our optimization jointly optimizes rigid-body motion, inertial parameters
and non-rigid scene deformation variables as defined in Eq.~\ref{eq:obs-vector-state-vio}.

As the window slides, new keyframes are added and the oldest ones are removed by marginalizing their corresponding states.
This marginalization step condenses past information into a compact prior term
$\mathcal{L}_{\mathrm{prior}}$, which preserves some degree of consistency while keeping
the computational complexity bounded.
The resulting objective function minimized by the system is given by
\begin{equation}
\mathcal{L}
=
\sum_{\tau \in \mathcal{W}}
\Big(
\mathcal{L}_{\mathrm{imu}}^{\tau}
+
\mathcal{L}_{\mathrm{rep}}^{\tau}
+
\lambda_{\mathrm{nr}} \, \mathcal{L}_{\mathrm{nr}}^{\tau}
\Big) \, + \mathcal{L}_{\mathrm{prior}},
\label{eq:full_cost}
\end{equation}
\noindent
where the scalar $\lambda_{\mathrm{nr}}$ balances the non-rigid expected motion against visual and inertial terms.

\subsection{Observability analysis}
Following~\cite{cerezoGNSS}, observability is assessed from a discrete-time perspective by
linearizing the measurement residuals over a finite temporal window.
Although Lie derivatives are not explicitly computed, the resulting Jacobians capture the
local sensitivity of the residuals with respect to the state, analogously to the
continuous-time formulation.
As shown in~\cite{hesch2014consistency,huang2009observability}, the rank of the stacked Jacobian
matrix determines the locally observable directions and is critical for estimator consistency.

Within this framework, we analyze observability over a two-keyframe window $\{\tau{-}1,\tau\}$,
jointly considering inertial preintegration, geometric visual, and non-rigid deformation
residuals acting on the state~\eqref{eq:obs-vector-state-vio}.
The corresponding observability matrix $\mathcal{O}$ is obtained by stacking the Jacobians of
all residuals with respect to the state vector $\boldsymbol{\xi}$, yielding
\begin{equation}
\label{eq:observability_matrix_vio}
\small
\begin{aligned}
\mathcal{O}
=\Big[
&\frac{\partial \mathbf{r}_{\Delta \R}^{\tau}}{\partial \boldsymbol{\xi}}^{\top},
\frac{\partial \mathbf{r}_{\Delta \mathbf{v}}^{\tau}}{\partial \boldsymbol{\xi}}^{\top},
\frac{\partial \mathbf{r}_{\Delta \mathbf{t}}^{\tau}}{\partial \boldsymbol{\xi}}^{\top} ,
\frac{\partial \mathbf{r}_{\mathrm{rep}}^{\tau}}{\partial \boldsymbol{\xi}}^{\top} ,
\frac{\partial \mathbf{r}_{\mathrm{elas}}^{\tau}}{\partial \boldsymbol{\xi}}^{\top} ,
\\
& \qquad \frac{\partial \mathbf{r}_{\mathrm{visc}}^{\tau}}{\partial \boldsymbol{\xi}}^{\top} ,
\frac{\partial \mathbf{r}_{\mathrm{photo}}^{\tau}}{\partial \boldsymbol{\xi}}^{\top} ,
\frac{\partial \mathbf{r}_{\mathbf{b}^g}^{\tau}}{\partial \boldsymbol{\xi}}^{\top} ,
\frac{\partial \mathbf{r}_{\mathbf{b}^a}^{\tau}}{\partial \boldsymbol{\xi}}^{\top} ,
\frac{\partial \mathbf{r}_{\mathbf{g}}^{\tau}}{\partial \boldsymbol{\xi}}^{\top}
\Big].
\end{aligned}
\end{equation}

\begin{table*}[t]
\centering
\scriptsize
\setlength{\tabcolsep}{3pt}
\renewcommand{\arraystretch}{1.15}

\caption{\textbf{Drunkard’s Dataset results.}
We report ATE RMSE, RPE, and number of successfully tracked frames. Each row averages all scenes at the same deformation level. Best result per metric in \textbf{bold}. Last row shows per-column means.}
\label{tab:drunkards-levels}
\rowcolors{4}{white}{gray!15}
\begin{tabular}{l l |
                c c |
                c c c |
                c c |
                c c c |
                c c |
                c c c}
\toprule
\multirow{3}{*}{Seq.} &
\multirow{3}{*}{Deformation} &

\multicolumn{5}{c|}{\textbf{ATE RMSE [mm]}} &
\multicolumn{5}{c|}{\textbf{RPE [mm]}} &
\multicolumn{5}{c}{\textbf{\#Frames}} \\

& &
\multirow{2}{*}{ORB-SLAM3} &
\multirow{2}{*}{NR-SLAM} &
\multicolumn{3}{c|}{\textbf{DefVINS}} &
\multirow{2}{*}{ORB-SLAM3} &
\multirow{2}{*}{NR-SLAM} &
\multicolumn{3}{c|}{\textbf{DefVINS}} &
\multirow{2}{*}{ORB-SLAM3} &
\multirow{2}{*}{NR-SLAM} &
\multicolumn{3}{c}{\textbf{DefVINS}} \\

& & & & {V-NR} & {VI-R} & {full}
  & & & {V-NR} & {VI-R} & {full}
  & & & {V-NR} & {VI-R} & {full} \\

\midrule
L0 & Low
 & 6.0  & \textbf{5.4}  & 9.2  & 7.1  & 6.8
 & 1.1  & \textbf{1.0}  & 2.2  & 1.9  & 1.2
 & 1987 & 2061 & 2124 & 2169 & \textbf{2198} \\

L1 & Medium
 & 19.4 & 11.6 & 17.1 & 13.2 & \textbf{9.4}
 & 2.1  & 2.0  & 3.1  & 2.4  & \textbf{2.0}
 & 1879 & 1968 & 2057 & 2096 & \textbf{2128} \\

L2 & Hard
 & 42.3 & 19.5 & 27.4 & 21.1 & \textbf{14.3}
 & 3.2  & \textbf{3.0}  & 4.1  & 3.4  & 3.1
 & 1746 & 1842 & 1928 & 1986 & \textbf{2021} \\

L3 & Extreme
 & 53.1 & 25.4 & 39.2 & 30.3 & \textbf{19.6}
 & 5.0  & 4.3  & 5.2  & 4.1  & \textbf{3.3}
 & 1612 & 1729 & 1814 & 1876 & \textbf{1919} \\
 
\midrule
Mean & 
 & 30.2 & 15.5 & 23.2 & 17.9 & \textbf{12.5}
 & 2.85 & 2.6  & 3.7  & 3.0  & \textbf{2.4}
 & 1806 & 1900 & 1981 & 2032 & \textbf{2067} \\
\bottomrule
\end{tabular}
\end{table*}

\color{black}

Based on the previously defined residuals, Fig.~\ref{fig:obs-matrix-explicit-vio} illustrates the
explicit block structure of the observability matrix $\mathcal{O}$.
For clarity and readability, the analytical expressions of the nonzero Jacobian blocks are
reported in the text rather than embedded in the figure.
Geometric reprojection residuals contribute with standard visual-inertial Jacobian blocks,
namely $\mathbf{H}_{\R_\tau}^{\mathrm{vis}} = -\mathbf{J}_\pi\,\mathbf{t}_{\mathrm{cam}}^\wedge$,
$\mathbf{H}_{\mathbf{t}_\tau}^{\mathrm{vis}} = -\mathbf{J}_\pi \R_t^\top$, and
$\mathbf{H}_{\mathrm{nr}}^{\mathrm{vis}} = \mathbf{J}_\pi \R_\tau^\top$ for the deformation nodes.
Non-rigid regularization terms introduce additional couplings among deformation variables:
the elastic prior yields
$\mathbf{H}_{\mathrm{nr}}^{\mathrm{elas}} \propto (\mathbf{x}_i^\tau - \mathbf{x}_j^\tau)/d_{ij}^\tau$,
the viscous term results in block structures of the form
$[\mathbf{I}, -\mathbf{I}, -\mathbf{I}, \mathbf{I}]$,
and the photometric term contributes
$\mathbf{H}_{\mathrm{nr}}^{\mathrm{photo}}
= \nabla I^\tau(\mathbf{u}_i^\tau)\,\mathbf{J}_\pi \R_\tau^\top$,
with analogous derivatives with respect to the pose variables.
Together, these expressions define all nonzero blocks of $\mathcal{O}$.

A symbolic inspection of the resulting Jacobian structure shows that the formulation preserves
the classical gauge freedoms of VIO while introducing additional degrees
of freedom associated with the non-rigid deformation field.
When the non-rigid variables $\{ \mathbf{x}_i^\tau \}_{i \in \mathcal{D},\tau \in \mathcal{W}}$ are ignored and metric depth is
assumed, the rigid subsystem remains observable only up to a global $SE(3)$ transformation:
absolute position and yaw are unobservable, whereas gravity magnitude and direction become
observable under sufficiently rich accelerations and rotations~\cite{hesch2014consistency}.
In particular, the IMU velocity and position residuals
$\mathbf{r}_{\Delta \mathbf{v}}^{\tau}$ and $\mathbf{r}_{\Delta \mathbf{t}}^{\tau}$
couple accelerometer bias and gravity, becoming rank-deficient under near-constant-velocity
motion, while insufficient rotational excitation leaves yaw and gyroscope bias coupled in
$\mathbf{r}_{\Delta \R}^{t}$.

The non-rigid substate is constrained by the combined action of
geometric reprojection, photometric consistency, and visco-elastic regularization.
Visual and photometric terms anchor node motion to image evidence, whereas elastic and viscous
priors regularize spatial and temporal variations, eliminating low-energy deformation modes
that would otherwise be consistent with visual measurements alone.
As a result, the rank of $\mathcal{O}$ increases in the non-rigid subspace.

Although inertial measurements do not directly observe the deformation field, they improve its
observability indirectly by stabilizing the global rigid trajectory.
Once gravity and inertial biases are sufficiently well conditioned, deformation nodes can no
longer absorb errors in global pose or yaw, reducing their ability to mimic rigid drift.
Consequently, sufficiently rich inertial excitation helps decouple rigid motion from non-rigid
deformation and further constrains the remaining deformation modes.

In practice, the rank of $\mathcal{O}$ depends on both motion and deformation patterns.
Trajectories with limited parallax or negligible rotation lead to weak observability of inertial and deformation states, whereas well-excited motions yield a well-conditioned system in which the rigid state is observable up to the
expected gauge freedoms and the deformation field is sufficiently constrained.


\section{Experiments}

\subsection{Experimental setup}
All experiments were conducted on a desktop PC with an Intel Core i7-11700K
(3.6\,GHz, 64\,GB RAM).
DefVINS is implemented in C++ and uses
Ceres\footnote{\url{http://ceres-solver.org/}} for optimization purposes.
Unless otherwise stated, all methods run in a single thread and wall-clock
times are reported.
Performance is evaluated using standard trajectory-based metrics.
We report Absolute Trajectory Error (ATE), Relative Position Error (RPE), and the number of successfully tracked frames as an indicator
of robustness.
Comparisons are performed against ORB-SLAM3~\cite{campos2021orb} as a rigid visual-inertial SLAM
baseline and NR-SLAM~\cite{gomez2024nrslam} as a non-rigid visual-only pipeline.

We first evaluate the method on the synthetic Drunkard’s Dataset~\cite{recasens2024drunkard}. The Drunkard’s Dataset provides 19 synthetic videos (320$\times$320) with full 3D ground truth
and four increasing deformation levels per scene, enabling controlled analysis. We generate IMU data by differentiating B-splines fitted to ground truth camera poses, ensuring temporally smooth and physically consistent signals.

We further validate DefVINS on our VIMandala dataset, comprising seven 848$\times$480 sequences under natural image noise, real sensor motion and uncontrolled surface deformations.
Synchronized RGB, depth, IMU, and ground-truth camera motion are recorded for the seven sequences. The sequences are denoted as R0--R6, where R0 corresponds to near-rigid motion and R6 represents the most severe deformations.
This gradual increase in deformation enables systematic evaluation under progressively more
challenging conditions. 

\subsection{Synthetic experiments in the Drunkard's dataset}


Table~\ref{tab:drunkards-levels} reports quantitative results on the Drunkard's sequences. Metrics are aggregated across scenes at each deformation level to provide compact yet representative results. For our DefVINS, we report results for the full pipeline, but also for the ablated versions V-NR (including deformation model but not inertial measurements) and VI-R (including inertial measurements but not deformation model).

\begin{table*}[ht!]
\centering
\scriptsize
\setlength{\tabcolsep}{3pt}
\renewcommand{\arraystretch}{1.20}

\caption{\textbf{Comparison of visual-inertial odometry methods on the VIMandala dataset.}
We report ATE RMSE, RPE, and number of successfully tracked frames. Best result per metric in \textbf{bold}. Last row shows per-column means.}
\label{tab:vio-unified}
\rowcolors{5}{gray!15}{white}
\begin{tabular}{l l |
                c c |
                c c c |
                c c |
                c c c |
                c c |
                c c c}
\toprule
\multirow{3}{*}{Seq.} &
\multirow{3}{*}{Deformation} &

\multicolumn{5}{c|}{\textbf{ATE RMSE [mm]}} &
\multicolumn{5}{c|}{\textbf{RPE [mm]}} &
\multicolumn{5}{c}{\textbf{\#Frames}} \\

& &
\multirow{2}{*}{ORB-SLAM3} &
\multirow{2}{*}{NR-SLAM} &
\multicolumn{3}{c|}{\textbf{DefVINS}} &
\multirow{2}{*}{ORB-SLAM3} &
\multirow{2}{*}{NR-SLAM} &
\multicolumn{3}{c|}{\textbf{DefVINS}} &
\multirow{2}{*}{ORB-SLAM3} &
\multirow{2}{*}{NR-SLAM} &
\multicolumn{3}{c}{\textbf{DefVINS}} \\

& & & & {V-NR} & {VI-R} & {full}
  & & & {V-NR} & {VI-R} & {full}
  & & & {V-NR} & {VI-R} & {full} \\
\midrule

R0 & Low
& \textbf{6.9} & 7.1
& 10.8 & 8.9 & 8.1
& \textbf{1.2} & 1.4
& 1.9 & 1.6 & 1.5
& \textbf{1810} & 1804
& 1805 & 1802 & 1804 \\

R1 & Low
& \textbf{7.5} & 8.6
& 13.2 & 10.6 & 9.0
& \textbf{1.4} & 1.6
& 2.1 & 1.8 & 1.6
& 1684 & 1743
& 1751 & 1743 & \textbf{1770} \\

R2 & Medium
& 15.3 & 10.5
& 19.6 & 15.1 & \textbf{10.2}
& 2.3 & {2.2}
& 2.7 & 2.5 & \textbf{2.0}
& 1589 & 1658
& 1706 & 1742 & \textbf{1768} \\

R3 & Medium
& 27.6 & 17.9
& 26.4 & 19.9 & \textbf{10.8}
& 3.5 & 3.0
& 3.1 & 2.5 & \textbf{2.1}
& 1496 & 1592
& 1651 & 1688 & \textbf{1719} \\

R4 & Medium
& 48.1 & 19.4
& 30.8 & 24.9 & \textbf{11.4}
& 4.6 & 3.4
& 3.1 & 2.0 & \textbf{1.9}
& 1387 & 1504
& 1568 & 1603 & \textbf{1736} \\

R5 & High
& 71.4 & 39.8
& 44.5 & 33.2 & \textbf{15.6}
& 6.1 & 4.5
& 4.0 & 2.9 & \textbf{2.4}
& 1194 & 1542
& 1623 & 1668 & \textbf{1706} \\

R6 & High
& 95.8 & 57.2
& 60.8 & 41.0 & \textbf{19.8}
& 7.8 & 5.2
& 4.7 & 3.5 & \textbf{3.0}
& 982 & 1476
& 1558 & 1604 & \textbf{1641} \\
\midrule
Mean & 
& 39.0 & 22.9
& 29.4 & 22.0 & \textbf{12.1}
& 3.84 & 3.04
& 3.09 & 2.40 & \textbf{2.07}
& 1449 & 1617
& 1666 & 1693 & \textbf{1735} \\

\bottomrule
\end{tabular}
\end{table*}

In low-deformation scenarios (L0), ORB-SLAM3, NR-SLAM and our DefVINS (full) achieve similarly low errors.
In this regime, the rigid visual-inertial baseline ORB-SLAM3 performs competitively against non-rigid approaches, indicating that explicit non-rigid modeling is not strictly necessary when deformations are minimal. Nevertheless, the incorporation of inertial measurements already provides measurable benefits: compared to the visual-only DefVINS configuration (V-NR), the rigid visual-inertial variant (VI-R) reduces ATE by roughly 20\%.
As deformation increases to moderate levels (L1), performance differences become more pronounced. While ORB-SLAM3 exhibits a substantial increase in ATE relative to L0, both NR-SLAM and DefVINS maintain significantly lower errors. In particular, our DefVINS (full) pipeline achieves an ATE reduction of approximately 30\% with respect to ORB-SLAM3 and nearly 45\% compared to the
V-NR baseline.

Improvements in RPE remain more moderate (around 20--30\%), suggesting that inertial constraints primarily contribute to stabilizing short-term motion estimates.
In harder deformation regimes (L2), rigid motion assumptions further degrade estimation accuracy.
Relative to ORB-SLAM3, the full DefVINS formulation reduces ATE by approximately 35--40\% and RPE by about 25\%.
Moreover, when compared to the rigid DefVINS variant (VI-R), the inclusion of explicit non-rigid regularization yields an additional ATE reduction of roughly 30\%, highlighting the importance of modeling deformation dynamics beyond inertial stabilization alone.
In the extreme deformation case (L3), the limitations of rigid models become most evident. The full DefVINS formulation achieves an ATE reduction of approximately 40--45\% with respect to ORB-SLAM3 and nearly 50\% compared to our visual-only V-NR configuration, while also reducing RPE by about 40\%. Although all methods experience a reduction in the number of successfully tracked frames as deformation severity increases, DefVINS consistently achieves the longest estimated trajectories, indicating improved robustness to severe
non-rigid motion. 

\begin{figure}
    \centering
    \includegraphics[width=\linewidth]{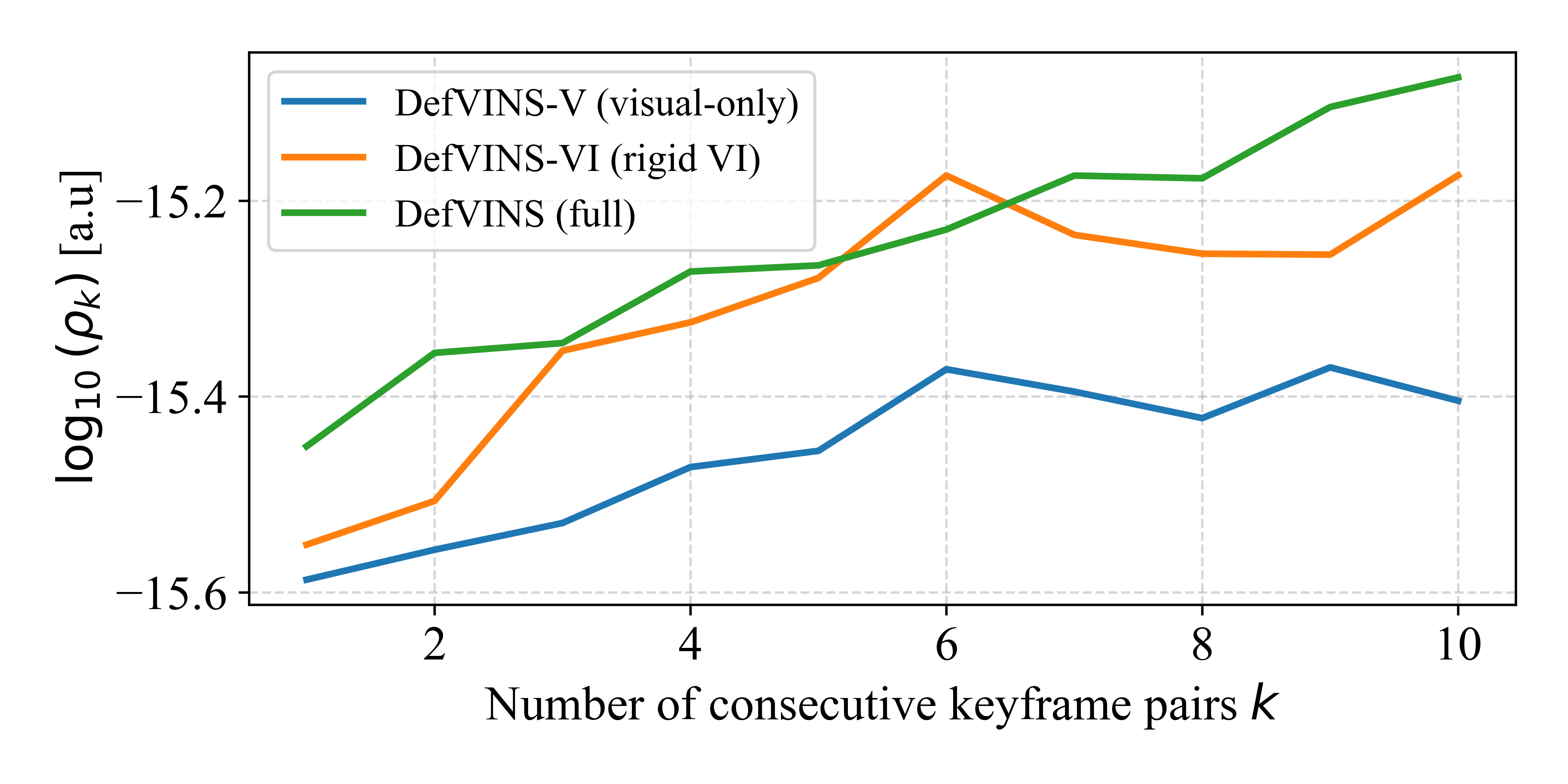}
    \caption{\textbf{Illustrative observability analysis under synthetic conditions.}
    Evolution of the conditioning score $\log_{10}(\rho_k)$ as a function of the
    number of stacked keyframe pairs $k$.
    Inertial sensing and
    non-rigid regularization significantly improve numerical conditioning,
    yielding well-observed directions with only a few frames.}
    \label{fig:obs-spectrum}
\end{figure}

Overall, results on the Drunkard's Dataset show that inertial sensing and explicit non-rigid modeling provide complementary benefits. While inertial constraints primarily improve local motion consistency, non-rigid modeling is essential to recover globally accurate trajectories as deformation amplitude increases, even under synthetic conditions.

\begin{figure*}[] 
  \centering
\includegraphics[width=2.0\columnwidth]{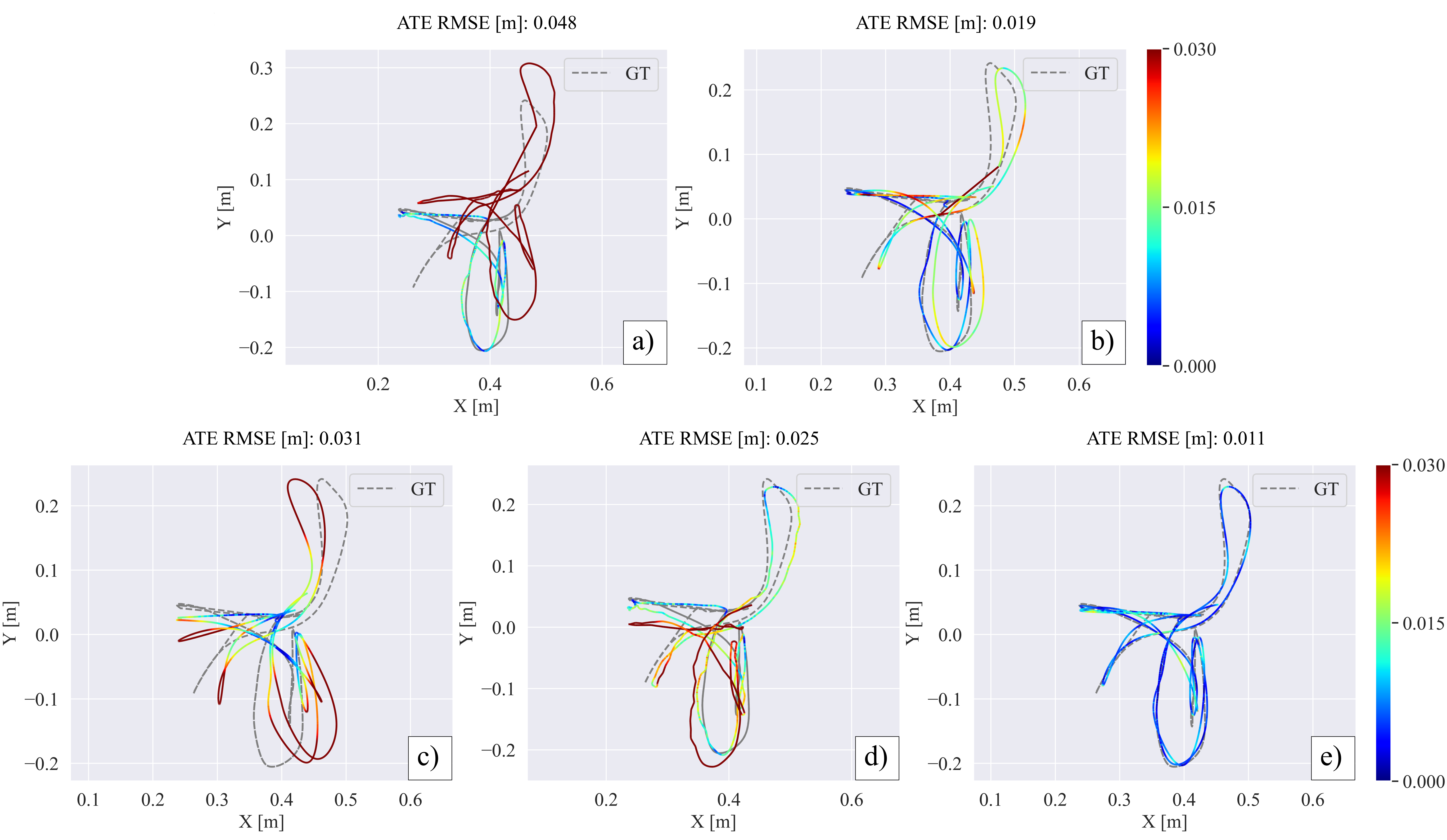}
\caption{\textbf{Comparison between DefVINS operating modes and baselines in VIMandala's R4 sequence.}
a) \textbf{ORB-SLAM3}: a representative rigid visual-inertial system. 
b) \textbf{NR-SLAM}: a representative non-rigid visual SLAM pipeline, with focus on medical applications.
c) \textbf{DefVINS-V} (visual-only, non-rigid): the lack of inertial constraints leads to accumulated drift, particularly during turning motions.
b) \textbf{DefVINS-VI} (rigid): introducing inertial sensing stabilizes rotational estimates, but assuming scene rigidity limits performance under deformation.
e) \textbf{DefVINS (full)}: by jointly enabling inertial sensing and explicit non-rigid modeling, DefVINS achieves the global consistency and lowest trajectory error. Dashed line represent the ground-truth while colored line
represent the camera trajectory and its estimation errors (see errors scale at the right).}
  \label{fig:trajectories-comparison}
\end{figure*}

\subsection{Observability analysis}
To complement the quantitative evaluation, we present an illustrative analysis of
numerical conditioning under controlled synthetic conditions. A set of short motion segments with rich 6-DoF excitation is selected from the VIMandala dataset. Specifically, we extract five non-overlapping segments of ten consecutive keyframes each, characterized by non-collinear translations and rotations about multiple axes. For each segment, residual Jacobians are generated and stacked over $k$ consecutive keyframe pairs to form the observability matrix $\mathcal{O}$. The reported results correspond to the average conditioning score across the
selected segments. Observability is quantified using the conditioning score $\rho_k = {\sigma_{\min}}/{\sigma_{\max}}$, where $\sigma_{\min}$ and $\sigma_{\max}$ denote the smallest and largest singular values of $\mathcal{O}$, respectively. Fig.~\ref{fig:obs-spectrum} shows $\log_{10}(\rho_k)$ as a function of the number of keyframe pairs for the visual-only, rigid visual-inertial, and full formulations. The numerical conditioning evolves very differently for the aforementioned configurations. Inertial constraints lift near-null modes associated with gravity and biases, and non-rigid regularization prevents deformation variables from absorbing rigid-body drift. As a result, the full formulation becomes better conditioned with only a few frames, explaining the improved robustness observed in the synthetic experiments.

\subsection{Real experiments in the VIMandala dataset}
Table~\ref{tab:vio-unified} summarizes the quantitative performance of all evaluated methods on the real deformable sequences R0--R6. Several consistent trends can be observed across deformation regimes.
For the nearly rigid sequences (R0--R1), the rigid visual-inertial baseline ORB-SLAM3 achieves the lowest ATE and RPE values, outperforming deformable VIO baselines by approximately 20--30\% in ATE. This confirms that classical rigid visual-inertial approaches remain highly
effective when surface deformation is small, by fitting their rigid model to parts of the scene that are quasi-rigid. We attribute ORB-SLAM's better results to a better engineered implementation. In this regime, all DefVINS variants are able to track almost the full sequence length, with only marginal accuracy degradation. As the degree of deformation increases to moderate levels (R2--R3), rigid assumptions progressively break down.
Although ORB-SLAM3 and NR-SLAM still maintain camera pose tracking, both exhibit a marked increase in ATE. 
In contrast, incorporating inertial measurements within the DefVINS framework (VI-R) reduces ATE by approximately 25--30\% with respect to the visual-only configuration (V-NR), while also lowering RPE by around 15--20\%. This highlights the stabilizing effect of inertial constraints on rotational estimation and short-term motion consistency. NR-SLAM consistently achieves intermediate performance, outperforming the rigid ORB-SLAM3 but still falling short of our proposed DefVINS. For medium-to-high deformation scenarios (R4--R6), the limitations of rigid
scene models become evident. Relative to ORB-SLAM3, the full DefVINS formulation reduces ATE by approximately 75\% on R4, around 80\% on R5, and close to 80\% on R6. At the same time, DefVINS (full) consistently tracks a substantially larger fraction of frames (around 85--95\%) compared to ORB-SLAM3, whose tracking coverage drops below 50\% in R4 and below 20\% in R6. While inertial sensing alone partially mitigates tracking degradation, only the joint combination of inertial constraints and an explicit model of non-rigidity preserves both accuracy and robustness under strong deformation. 

Overall, these results demonstrate that inertial sensing and non-rigid modeling play complementary roles. Inertial data primarily stabilize local motion and reduce short-term drift, whereas explicit non-rigid regularization is essential to maintain global consistency and long-term tracking in highly deformable environments. It is also worth highlighting that DefVINS improvements over baselines, already significant in the synthetic Drundard's data, become even more evident in the real setups of our VIMandala sequences. 

\begin{figure}[]
    \centering
    \includegraphics[width=0.99\linewidth]{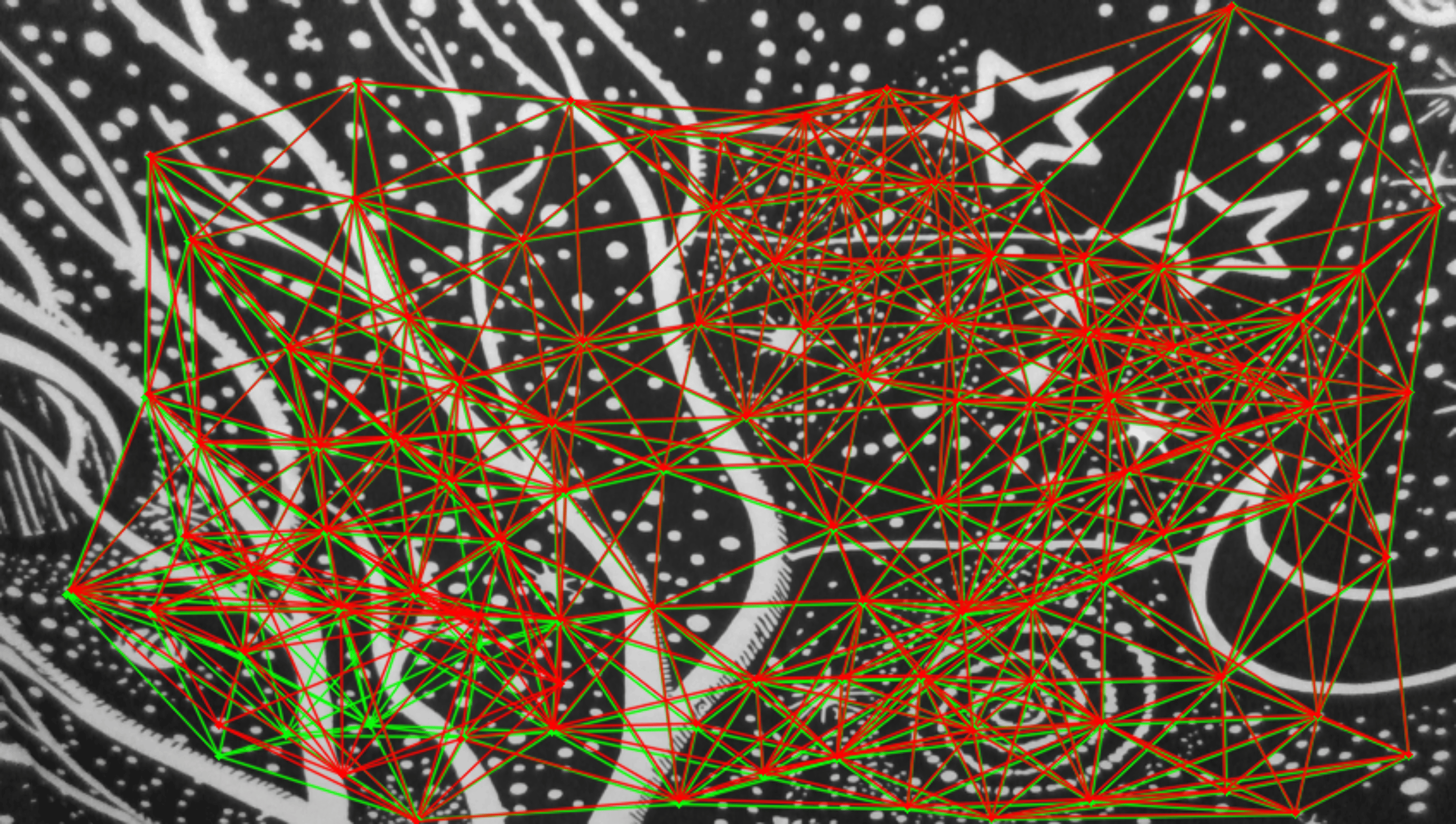}
\caption{\textbf{Deformation graph on sequence R4.}
Green and red edges denote the deformation graph for consecutive frames at times $t_\tau$ and $t_{\tau+1}$, respectively.
Their differences indicate medium-to-high non-rigid deformation, with stronger effects in the lower-left region.}
    \label{fig:mandala-deformation-graph}
\end{figure}

To assess the contribution of each system component, the proposed framework is evaluated
under multiple operating modes obtained by selectively enabling inertial sensing and
non-rigid modeling.
Fig.~\ref{fig:trajectories-comparison} qualitatively compares the resulting camera trajectories
for a visual-only configuration (DefVINS-V), a rigid visual-inertial setup without
non-rigid regularization (DefVINS-VI), and the full DefVINS model, which jointly exploits
inertial constraints and non-rigid motion modeling.
Results are also compared against ORB-SLAM3 and NR-SLAM.
This ablation study highlights the complementary roles of inertial sensing and non-rigid
regularization, and exposes the limitations of rigid motion assumptions in deformable scenes.

Fig.~\ref{fig:mandala-deformation-graph}
illustrates the estimated deformation graph for a representative image of our VIMandala dataset.
The graph reveals spatially varying deformation, with larger displacements in the lower-left
region and, comparatively, quasi-rigid behavior elsewhere, demonstrating the ability of the proposed
model to capture localized deformations while maintaining global
surface coherence.

\section{Conclusions}

\noindent This paper introduced DefVINS, a visual-inertial odometry
pipeline for deformable scenes that decouples a rigid, IMU-anchored state from non-rigid scene deformation parameters. In addition, given the lack of benchmarks for this task, we contribute by synthesizing IMU readings for the synthetic Drunkard's dataset and curating a new real benchmark imaging a deforming mandala.
Our experiments on both synthetic and real data demonstrate that DefVINS provides accurate and stable state estimation across a wide
range of deformation regimes, outperforming relevant rigid visual-inertial and non-rigid visual baselines.

\balance

\end{document}